\documentclass[10pt,twocolumn,letterpaper]{article}

\usepackage{cvpr}              % To produce the CAMERA-READY version
\usepackage{multirow}

\usepackage{booktabs}   % 三线表基础
\usepackage{xcolor}    % 颜色支持
\usepackage{colortbl}  % 表格线条染色（arrayrulecolor）
\definecolor{cvprblue}{rgb}{0.21,0.49,0.74}
\usepackage[pagebackref,breaklinks,colorlinks,allcolors=cvprblue]{hyperref}

\def\paperID{*****} % *** Enter the Paper ID here
\def\confName{CVPR}
\def\confYear{2026}

\title{DrivingDepth: Sparse-Prompted Pixel-wise Scale Correction for \\ 
Driving Depth Estimation}
\author{
\textbf{Chi Huang}$^{\dagger}$\quad
\textbf{Wenhao Zhang}$^{}$\quad
\textbf{Hang Yin}$^{}$\quad
\textbf{YuAn Wang}$^{}$\quad\\
\textbf{Hao Li}$^{}$\thanks{Team Lead}\quad
\textbf{Bosheng Wang}$^{}$\quad
\textbf{Xun Sun}$^{}$\quad
\textbf{Liang Wang}$^{}$\quad
\\[2pt]
$^{}$Baidu Inc.\\[2pt]
}

\begin{document}
% \maketitle

\twocolumn[{%
\maketitle
 \vspace{-35pt}
\begin{center}
  \includegraphics[width=0.9\textwidth]{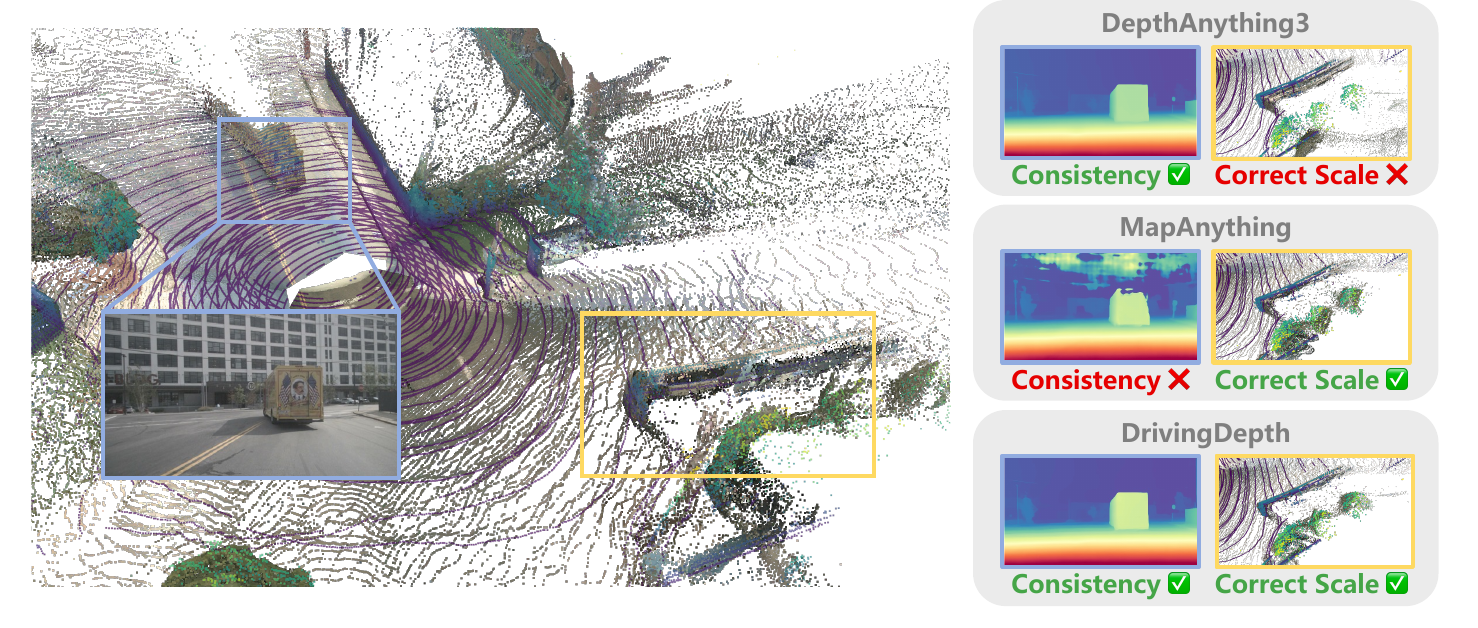}
   \vspace{-10pt}
  \captionof{figure}{\textbf{Comparison of \emph{DrivingDepth} with DepthAnything3~\cite{depthanything3} and MapAnything~\cite{mapanything}.} \textbf{Left:} DrivingDepth's full-scene 3D reconstruction. \textbf{Right:} 2D depth maps and 3D point clouds. 
  % LiDAR map overlaid on point cloud. 
  LiDAR points colored by Z-height; point clouds with projected image RGB.
  DrivingDepth maintains RGB-depth consistency while yielding correct scales.
  More demos are available at  \href{https://hcaelrs.github.io/DrivingDepth-page/}{\emph{DrivingDepth-page}}.
}
  \label{fig:teaser}
\end{center}
}]
% \renewcommand{\thefootnote}{\fnsymbol{footnote}}%
% \footnotetext[2]{Team Lead}%
% \renewcommand{\thefootnote}{\arabic{footnote}}%
\vspace{-10pt}

\begin{abstract}
% \vspace*{-20pt}\noindent
\vskip -10pt
Dense depth estimation for autonomous driving faces a \emph{geometry--scale conflict}: depth foundation models deliver pixel-aligned dense visual geometry without reliable metric scale, while projected LiDAR provides metric anchors that are sparse, noisy, and misaligned with image structures.
Existing sparse-prompted methods incorporate LiDAR by regenerating depth from scratch, overriding the foundation model's coherent geometry and producing structural artifacts on visually continuous surfaces.
% We observe that foundation models already capture geometrically coherent relative depth; what remains to be learned is not surface structure but only a per-pixel scale factor mapping relative geometry to metric coordinates.
Our key insight is that foundation models already capture geometrically coherent relative depth; no additional surface structure learning is required—only a per-pixel scale factor mapping relative geometry to metric coordinates.
Based on this, we propose DrivingDepth, which treats sparse LiDAR as \emph{geometric prompts} that locally calibrate a frozen foundation prior through residual pixel-wise scale correction, preserving dense visual geometry by construction.
% A Geometry-Preserving Feature Adapter injects sparse-depth cues into the frozen backbone via cross-attention and propagates them along co-visible frame--view connections, while surface-normal regularization ensures that the learned correction maintains image-aligned structure between sparse anchors.
% On nuScenes with 4-frame surround-view 6-camera input, DrivingDepth achieves an AbsRel of 11.19 and an EdgeCR of 5.741, compared to 11.99 and 1.914 for MapAnything.
On nuScenes with 4-frame surround-view input, DrivingDepth achieves an AbsRel of 11.19 and an EdgeCR of 5.741, outperforming MapAnything (11.99/1.914) by simultaneously delivering SOTA metric accuracy and geometric consistency.
\end{abstract}
% \vspace*{-25pt}
% Dense depth estimation for autonomous driving faces a \emph{geometry--scale conflict}: depth foundation models provide dense visual geometry without reliable metric scale, while projected LiDAR provides metric anchors that are sparse, noisy, and misaligned with image structures.
% Existing sparse-prompted methods fuse LiDAR by regenerating depth from scratch, overriding the foundation model's coherent geometry and producing structural artifacts on visually continuous surfaces.
% We observe that foundation models already capture geometrically coherent relative depth; what remains to be learned is not surface structure but only a per-pixel scale factor mapping relative geometry to metric coordinates.
% Based on this observation, we propose DrivingDepth, which treats sparse LiDAR as \emph{geometric prompts} that locally calibrate a frozen foundation prior through residual pixel-wise scale correction, preserving dense visual geometry by construction.
% On nuScenes with 4-frame surround-view 6-camera input, DrivingDepth achieves an AbsRel of 11.19 and an EdgeCR of 5.741, compared to 11.99 and 1.914 for MapAnything.

\vspace{-20pt}
% \vskip -50pt
%======================================================================
\section{Introduction}
%======================================================================
\vskip -3pt

Autonomous driving demands depth estimation that is simultaneously dense, pixel-aligned, and metrically consistent---yet no single sensing modality fulfills all three requirements.
Downstream tasks such as 3D reconstruction and occupancy prediction rely on pixel-aligned depth: edge misalignment yields floaters and distorted voxel boundaries.
Cameras capture rich semantic layout without measuring absolute distance; LiDAR yields metric measurements that become sparse, unevenly distributed, and often misaligned with image boundaries upon projection onto the image plane.
Calibration drift, temporal offsets, and surface reflections further degrade these projections, making sparse metric evidence unreliable as dense ground truth.
We term this tension the \emph{geometry--scale conflict}: dense visual geometry and reliable metric scale originate from fundamentally different sources, and naively combining them risks corrupting one to obtain the other.

Existing methods each address only part of this problem.
Monocular depth estimation models such as DepthAnything~\cite{depthanything}, Depth Pro~\cite{depthpro}, and MOGE-2~\cite{moge2} produce strong dense visual geometry but reason within each image independently.
% End-to-end renconstruction models~\cite{depthanything3, vggt,dust3r,mapanything} recover coherent 3D structure from correspondences but are not designed to incorporate projected LiDAR for metric calibration.
Visual geometry grounded foundation models such as VGGT~\cite{vggt}, DUSt3R~\cite{dust3r}, and DepthAnything3~\cite{depthanything3} exploit correspondences across multiple input images to recover coherent 3D structure, but are not designed to incorporate projected LiDAR for metric calibration.
Sparse-prompted methods such as PromptDA~\cite{promptda}, PriorDA~\cite{prioranything}, and MapAnything~\cite{mapanything} can incorporate LiDAR-like inputs, but they predict absolute depth from fused features rather than correcting an existing geometric prior, allowing the model to freely deviate from the foundation's coherent structure wherever LiDAR evidence is sparse or noisy---producing geometric inconsistencies such as holes or broken surfaces on visually continuous objects (Figure~\ref{fig:teaser}).
These limitations illustrate the geometry--scale conflict in practice: LiDAR-guided depth prediction compromises depth-image consistency, while the visual foundation model alone lacks metric scale.

A global post-hoc alignment can adjust average scale but cannot correct spatially varying metric errors across different surfaces and depth ranges.
The goal is therefore to let sparse metric evidence anchor scale at reliable locations through pixel-wise correction, while preserving the dense visual geometry between anchors.
This decomposition is feasible because foundation models already produce geometrically coherent relative depth; what remains to be learned is not surface structure but only the per-pixel scale that maps relative geometry to metric coordinates. 

To this end, we propose \emph{DrivingDepth}, a sparse-prompted dense depth framework built on DepthAnything3 (DA3)~\cite{depthanything3} that follows a minimal-intervention principle: the model is initialized to reproduce the original DA3~\cite{depthanything3} output and learns only a residual pixel-wise scale correction for metric calibration, preserving the foundation geometry by design.
% low cost: emphasize that the minimal-intervention design keeps training affordable on a single 8-GPU node.
% As only a lightweight scale head and a thin feature adapter are trainable while the foundation backbone remains frozen, the entire framework can be trained on a single 8-GPU node, in contrast to end-to-end sparse-prompted alternatives that retrain dense depth predictors at substantially higher cost.
As the foundation model is frozen and only the lightweight scale head and feature adapter are learnable, our model fits one 8-GPU node, outperforming end-to-end sparse-prompt baselines with costly dense depth retraining.
Projected LiDAR is treated as \emph{sparse geometric prompts}---metric anchors that locally calibrate this dense visual geometry rather than dense supervision targets.
The framework operates at two complementary levels: at the feature level, a \emph{Geometry-Preserving Feature Adapter} injects sparse-depth cues into the frozen DA3~\cite{depthanything3} representation via cross-attention and propagates them through constrained self-attention restricted to adjacent cameras at the same timestamp and the same camera across all frames; at the output level, a \emph{Sparse-Aware Pixel-Scale Head} predicts per-pixel corrections together with a learned confidence map that automatically downweights unreliable LiDAR projections.
Surface-normal regularization further prevents the correction from naively fitting noisy points and preserves plausible surfaces between sparse anchors. We summarize our main contributions as follows: 
\begin{enumerate}[leftmargin=*]
    \item We identify the geometry--scale conflict in driving depth estimation and propose to resolve it through residual pixel-wise scale correction of a frozen foundation prior, where sparse LiDAR serves as geometric prompts for metric calibration rather than dense supervision targets.
    \item We design a minimal-intervention architecture that operates at two levels: a Geometry-Preserving Feature Adapter injects sparse-depth cues into frozen backbone features along co-visible frame--view connections, while a Sparse-Aware Pixel-Scale Head with learned confidence performs output-level correction.
    \item We propose a training objective that explicitly balances sparse metric anchoring against dense geometry preservation through confidence-weighted alignment, surface-normal regularization, and scale smoothness.
    % \item Experiments on nuScenes~\cite{nuscenes} and DDAD~\cite{ddad}---including sparse-prompt robustness under 10\% LiDAR density and surface-normal regularization ablation---demonstrate that DrivingDepth achieves competitive metric accuracy while maintaining strong boundary consistency.
    \item Experiments on nuScenes~\cite{nuscenes} and DDAD~\cite{ddad}, including sparse-prompt robustness under 10\% LiDAR density and surface-normal regularization ablation, demonstrate that DrivingDepth achieves competitive metric accuracy while maintaining strong geometric consistency.

\end{enumerate}

%======================================================================
\section{Related Work}
%======================================================================
\begin{figure*}[t]
  \centering
  \includegraphics[width=\textwidth]{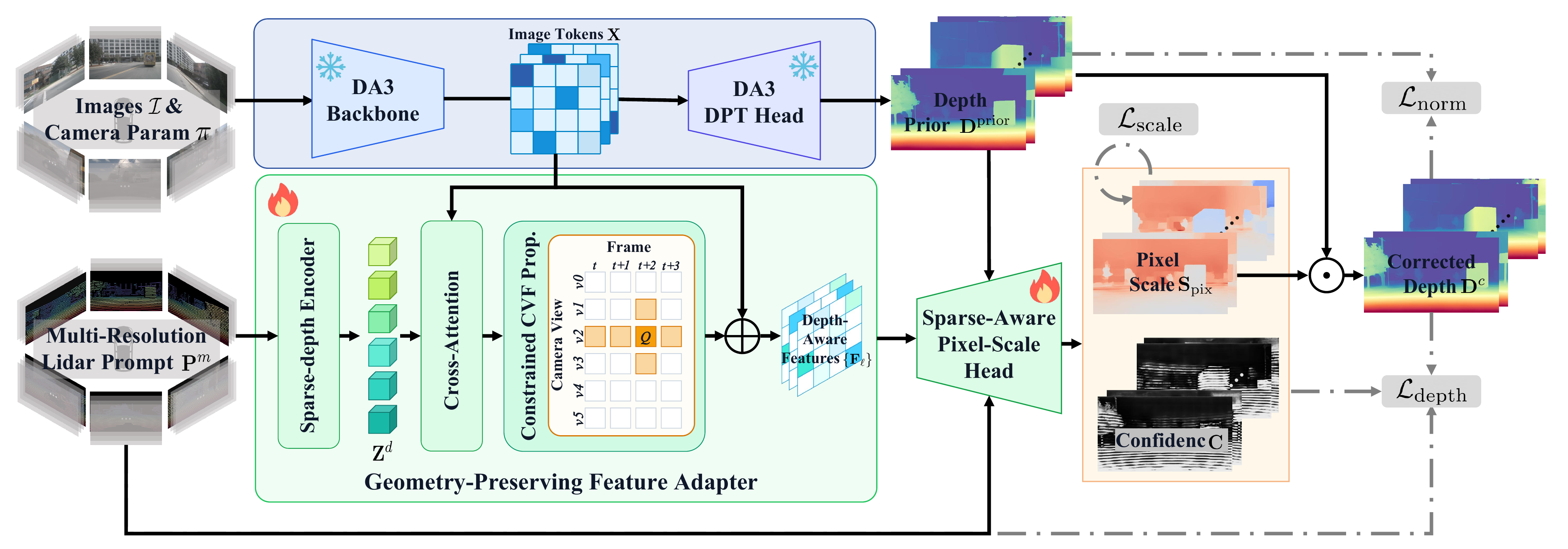}
  \vspace{-15pt}
  \caption{\textbf{Architecture of DrivingDepth.} The frozen DA3~\cite{depthanything3} maps surround-view images $\mathcal{I}$ and camera parameters $\pi$ to a dense depth prior $\mathbf{D}^{\mathrm{prior}}$. The Geometry-Preserving Feature Adapter fuses image tokens with sparse-depth tokens and propagates them under constrained cross-view/frame attention, yielding depth-aware features $\{\mathbf{F}_{\ell}\}$. The Sparse-Aware Pixel-Scale Head turns $\{\mathbf{F}_{\ell}\}$ and a LiDAR prompt into a per-pixel correction $\mathbf{S}_{\mathrm{pix}}$, which calibrates $\mathbf{D}^{\mathrm{prior}}$ into the corrected depth ${\mathbf{D}}^{c}$.}
  \label{fig:structure}
  \vspace{-10pt}
\end{figure*}

\subsection{Depth Estimation}

Monocular depth estimation has evolved from early paradigms---including continuous supervised regression~\cite{depthmap, depthlearning, bts}, discretized prediction~\cite{cao2017estimating, DORN, Adabins}, and global geometric reasoning~\cite{dpt, CRF, monodepth,monodepth2}---to recent foundation models~\cite{depthanything, depthpro, moge2, unidepth, metric3dv2, infinidepth} that produce robust dense visual geometry. While these models achieve robust generalization and exceptional boundary quality~\cite{depthanything, yang2024depthv2, depthpro, moge2}, they inherently lack reliable metric calibration. To address this in autonomous driving, methods exploit multi-view constraints~\cite{ddad, surrounddepth, m2depth} or incorporate projected LiDAR to anchor the metric space~\cite{promptda, prioranything, mapanything, completionformer}. However, sparse-guided approaches like MapAnything~\cite{mapanything} and PriorDA~\cite{prioranything} regenerate depth from scratch, often overwriting the foundation model's coherent surfaces and producing structural artifacts where LiDAR is sparse or noisy. In contrast, DrivingDepth treats sparse LiDAR strictly as metric prompts for the residual calibration of a frozen dense prior, bridging the geometry--scale conflict while preserving continuous visual structures by design.

\subsection{Feed-Forward 3D Reconstruction}

Feed-forward 3D reconstruction~\cite{dust3r,cut3r,depthanything3,mapanything,vggt,pi3,ttt3r,scal3r,fastvggt,dist4d} infers camera geometry, dense depth, correspondence, and scene structure directly from image collections without per-scene optimization.
DUSt3R~\cite{dust3r} and MASt3R~\cite{mast3r} use point-map representations to couple image matching with 3D recovery, while VGGT~\cite{vggt} and CUT3R~\cite{cut3r} broaden this paradigm toward unified multi-view prediction and continuous 3D perception.
DepthAnything3~\cite{depthanything3} further recovers spatially consistent visual geometry from arbitrary views, and MapAnything~\cite{mapanything} extends feed-forward reconstruction toward metric 3D outputs.
Although these models provide strong geometric priors or metric reconstruction baselines, they either rely mainly on visual correspondences or regenerate metric geometry from fused inputs.
DrivingDepth targets a narrower operation for driving depth: it keeps the feed-forward prior as a dense scaffold and learns a per-pixel multiplicative scale correction from sparse LiDAR prompts.

\subsection{Feed-Forward Driving Scene Reconstruction}
% \subsection{Feed-Forward Reconstruction in Autonomous Driving}
Feed-forward driving scene reconstruction~\cite{dggt,dvgt,drivingforward,vgd,streetforward,dynamicvggt} is particularly promising for autonomous driving because calibrated surround-camera streams provide structured multi-view observations at scale and require efficient 3D scene understanding.
% Feed-forward reconstruction is particularly promising for autonomous driving because calibrated surround-camera streams provide structured multi-view observations at scale and require efficient 3D scene understanding.
DVGT~\cite{dvgt} targets driving visual geometry, DGGT~\cite{dggt} reconstructs dynamic 4D scenes from unposed images, and DrivingForward~\cite{drivingforward} and VGD~\cite{vgd} use Gaussian representations for surround-view reconstruction and rendering.
These works demonstrate the potential of feed-forward reconstruction for driving geometry, but focus on scene reconstruction, novel-view synthesis, or 4D modeling; DrivingDepth instead isolates sparse-prompted metric depth calibration by correcting the scale of a frozen DA3 prior while preserving dense image-aligned geometry.

%======================================================================
\section{Method}
%======================================================================

\subsection{Overview}
\label{sec:method_overview}

% Given surround-view images $\mathcal{I} = \{\mathbf{I}_{t,v}\}$ across $T$ frames and $V$ cameras together with projected sparse metric LiDAR depth $\mathbf{D}^{\mathrm{sp}}$, the frozen foundation model $f_{\theta}$ (DA3) provides a dense depth prior $\mathbf{D}^{\mathrm{prior}} = f_{\theta}(\mathcal{I})$ with coherent surface geometry but no reliable metric scale.
% Because the residual scale error varies across surfaces and depth ranges, no single global factor can recover metric depth.
% DrivingDepth therefore keeps $\mathbf{D}^{\mathrm{prior}}$ as the dense visual geometry and learns a pixel-wise scale correction on top of it.
Given surround-view images $\mathcal{I} = \{\mathbf{I}_{t,v}\}$ across $T$ frames and $V$ cameras together with their camera parameters $\pi$
% and projected sparse metric LiDAR depth $\mathbf{D}^{\mathrm{sp}}$
, the frozen foundation model $f_{\theta}$ (DA3~\cite{depthanything3}) provides a dense depth prior $\mathbf{D}^{\mathrm{prior}} = f_{\theta}(\mathcal{I}, \pi)$ with coherent surface geometry but no reliable metric scale.
Because the residual scale error varies across surfaces and depth ranges, no single global factor can recover metric depth.
DrivingDepth therefore keeps $\mathbf{D}^{\mathrm{prior}}$ as the dense visual geometry and learns a pixel-wise scale correction on top of it.

We decompose this global metric correction into a per-pixel scale map $\mathbf{S}_{\mathrm{pix}}$ and a clip-level scalar $s_{\mathrm{g}}$, which turn the prior into the final metric depth $\hat{\mathbf{D}}^{c}$ via an intermediate locally-corrected depth $\mathbf{D}^{c}$:
\begin{equation}
    \mathbf{D}^{c} = \mathbf{D}^{\mathrm{prior}} \odot \mathbf{S}_{\mathrm{pix}}, \qquad
    \hat{\mathbf{D}}^{c} = \mathbf{D}^{c} \cdot s_{\mathrm{g}},
    \label{eq:scale_decomposition}
\end{equation}
where $\odot$ denotes element-wise multiplication and $s_{\mathrm{g}}$ is shared across all frames and views in a clip to keep the resulting point clouds metrically consistent.
We obtain $s_{\mathrm{g}}$ by ROE~\cite{moge1} alignment of $\mathbf{D}^{c}$ to the projected sparse metric LiDAR depth $\mathbf{D}^{\mathrm{sp}}$,
% \vspace{-5pt}
\begin{equation}
    s_{\mathrm{g}} = \mathrm{ROE}(\mathbf{D}^{c},\,\mathbf{D}^{\mathrm{sp}}),
    \label{eq:sglobal}
\end{equation}
$s_{\mathrm{g}}$ does not enter back-propagation.
The corrected depth $\hat{\mathbf{D}}^{c}$ is the final output of our method.
Unlike monocular methods that conduct independent frame-wise alignment, we compute a single $s_{\mathrm{g}}$ shared across all frames and views within a clip, ensuring metrically consistent point clouds across the entire surround-view sequence. 

We learn $\mathbf{S}_{\mathrm{pix}}$ at two complementary levels.
At the feature level, a \emph{Geometry-Preserving Feature Adapter} (Section~\ref{sec:pix_scale_feat}) injects sparse-depth cues into the frozen DA3 backbone along co-visible frame--view connections, producing depth-aware features $\{\mathbf{F}_{\ell}\}$.
At the output level, a \emph{Sparse-Aware Pixel-Scale Head} (Section~\ref{sec:pix_scale_head}) consumes the extracted $\{\mathbf{F}_{\ell}\}$ together with a multi-resolution LiDAR prompt to predict $\mathbf{S}_{\mathrm{pix}}$.
We further adopt a balanced training objective (Section~\ref{sec:training_loss}) that mediates between metric anchoring and dense geometry preservation.

\subsection{Geometry-Preserving Feature Adapter}
\label{sec:pix_scale_feat}

The adapter injects sparse-depth cues into the frozen DA3 representation while keeping its geometry intact in two senses: the backbone weights stay frozen, so DA3's dense visual geometry is never overwritten; and the depth-aware features are \emph{concatenated} with rather than replacing the image tokens, so the downstream scale head always retains direct access to the unmodified backbone.
All newly introduced branches are initialized neutrally so the model reproduces the original DA3 prediction at the start of training and learns corrections progressively.

\paragraph{Sparse-depth tokenization.}
The adapter is inserted at every output layer of the frozen backbone.
Let $\mathbf{X}_{\ell}\in\mathbb{R}^{B\times TV\times N\times C}$ denote the image tokens at layer $\ell$, with batch size $B$, $TV$ frame--view pairs, $N$ patch tokens per view, and $C$ channels.
For each $(t,v)$, a lightweight CNN with a patch-embedding layer $E_{d}$ encodes the per-view multi-resolution LiDAR prompt $\mathbf{P}^{m}$ (constructed in Section~\ref{sec:pix_scale_head}) independently into sparse-depth tokens $\mathbf{Z}^{d}=E_{d}(\mathbf{P}^{m})$ that share the patch grid of $\mathbf{X}_{\ell}$, and these tokens condition the image tokens via cross-attention:
\begin{equation}
    \tilde{\mathbf{X}}_{\ell}=\mathrm{Attn}_{\mathrm{cross}}(Q{=}\mathbf{X}_{\ell},\,K{=}V{=}\mathbf{Z}^{d}).
\end{equation}

% \paragraph{Constrained cross-view/frame propagation.}
% After cross-attention, sparse-depth evidence still needs to propagate across the $TV$ token sequence, but most camera pairs in a surround-view rig share no field of view: front and rear views observe disjoint surfaces, so unconstrained self-attention between them captures only spurious correlations and risks transferring scale evidence across surfaces that do not co-exist.
% We therefore restrict self-attention to two co-visible patterns---\emph{adjacent cameras at the same timestamp}, which share visual overlap, and \emph{the same camera across all frames in the clip}, which observes the same scene as the ego vehicle moves---yielding the depth-aware tokens $\mathbf{X}^{d}_{\ell}=\mathrm{Attn}_{\mathrm{constrained}}(\tilde{\mathbf{X}}_{\ell})$.
% The resulting attention connectivity forms a sparse $T{\times}V$ pattern that exposes only geometrically meaningful pairs.
% Finally, we concatenate the unmodified backbone features with the depth-aware adapted features as $\mathbf{F}_{\ell}=\mathrm{cat}(\mathbf{X}_{\ell},\mathbf{X}^{d}_{\ell})$, so the scale head in Section~\ref{sec:pix_scale_head} sees both the original geometric prior and the LiDAR-conditioned cues.

\paragraph{Constrained cross-view/frame propagation.}
After cross-attention, sparse-depth evidence still needs to propagate across the $TV$ token sequence. In a surround-view driving camera setup, however, only certain camera pairs share a field of view: adjacent cameras at the same timestamp overlap on shared surfaces, and the same camera across consecutive frames observes the same scene under ego motion, while non-adjacent views (e.g., front and rear) capture disjoint regions of the world. We therefore restrict self-attention to these two geometrically motivated patterns---\emph{adjacent cameras at the same timestamp} and \emph{the same camera across all frames in the clip}---yielding the depth-aware tokens $\mathbf{X}^{d}_{\ell}=\mathrm{Attn}_{\mathrm{constrained}}(\tilde{\mathbf{X}}_{\ell})$. The resulting attention connectivity forms a sparse $T{\times}V$ pattern aligned with the setup's co-visibility structure. Finally, we concatenate the unmodified backbone features with the depth-aware adapted features as $\mathbf{F}_{\ell}=\mathrm{cat}(\mathbf{X}_{\ell},\mathbf{X}^{d}_{\ell})$, so the scale head in Section~\ref{sec:pix_scale_head} sees both the original geometric prior and the LiDAR-conditioned cues.

\subsection{Sparse-Aware Pixel-Scale Head}
\label{sec:pix_scale_head}

Given the depth-aware features extracted from the feature adapter, the scale head produces the per-pixel scale map $\mathbf{S}_{\mathrm{pix}}$ that residually calibrates the DA3~\cite{depthanything3} prior in Eq.~\ref{eq:scale_decomposition}.
Because the residual scale error varies spatially across surfaces and depth ranges, a single global factor cannot fully resolve it; 
% we therefore predict a dense scale map rather than a scalar, and attach a PromptDA-DPT head repurposed to predict $\mathbf{S}_{\mathrm{pix}}$ rather than dense depth directly.
we therefore predict a dense scale map rather than a scalar, and attach a DPT head similar to that in PromptDA~\cite{promptda}, which is repurposed to predict $\mathbf{S}_{\mathrm{pix}}$ rather than dense depth directly.

Projected driving LiDAR is highly sparse---at native resolution, most decoder feature maps contain very few valid prompts with significant noise.
We construct a multi-resolution LiDAR prompt $\mathbf{P}^{m}$ by downsampling sparse depth to multiple levels and upsampling back, increasing effective prompt density for lower-resolution decoder stages.
Together with the DA3 prior, the LiDAR prompt is:
\begin{align}
    \mathbf{P}^{m}&=\mathrm{cat}_{l}\bigl([\hat{\mathbf{D}}^{\mathrm{sp}}_{l},\hat{\mathbf{M}}^{\mathrm{sp}}_{l}]\bigr), \\
    \mathbf{P}^{L}&=\mathrm{cat}(\mathbf{P}^{m},\;\mathbf{D}^{\mathrm{prior}}),
\end{align}
where $l$ indexes the resolution level, $\hat{\mathbf{D}}^{\mathrm{sp}}_{l}$ is the sparse depth downsampled to level $l$, upsampled back to native resolution, and normalized to a relative scale, $\hat{\mathbf{M}}^{\mathrm{sp}}_{l}$ is the corresponding validity mask at level $l$, $\mathrm{cat}_{l}$ concatenates all levels along the channel dimension, and $\mathbf{P}^{L}$ is the complete LiDAR prompt.
By concatenating  $\mathbf{D}^{\mathrm{prior}}$ into $\mathbf{P}^{L}$ 
% enables the head to contrast sparse cues against the estimate it is correcting.
, the head can compare sparse LiDAR signals against the existing reliable depth estimate and further refine it.

A PromptDA-style~\cite{promptda} decoder $H_{\mathrm{scale}}$ injects $\mathbf{P}^{L}$ through zero-initialized convolutions---ensuring the LiDAR cues contribute nothing at initialization so the model starts from unmodified backbone features---into the multi-level features $\{\mathbf{F}_{\ell}\}$ produced by the adapter, and outputs a scale map together with a confidence map:
\begin{equation}
    [\mathbf{S}_{\mathrm{pix}},\mathbf{C}]=H_{\mathrm{scale}}(\{\mathbf{F}_{\ell}\},\mathbf{P}^{L}).
    \label{eq:spix}
\end{equation}
Letting $x$ and $c$ denote the raw scale and confidence logits directly produced by the decoder, the final per-pixel scale is parameterized as:
\begin{equation}
    \mathbf{S}_{\mathrm{pix}} = \exp\bigl(\alpha\cdot(2\,\sigma(x)-1)\bigr),
    \label{eq:scale_param}
\end{equation}
where $\sigma(\cdot)$ denotes the sigmoid function and $\alpha$ is a learnable coefficient initialized to $0.5$.
The $\exp(\cdot)$ parameterization ensures symmetric scaling in log-space, avoiding the inherent asymmetry of multiplicative correction in linear parameterization; at initialization $\sigma(0){=}0.5$ yields $\mathbf{S}_{\mathrm{pix}}{=}1$ everywhere, so the model begins from the unmodified DA3 prediction and learns corrections progressively.
The confidence map $\mathbf{C}=\sigma(c)\in(0,1)$ is used in the training loss to downweight unreliable LiDAR projections.

\subsection{Training Loss}
\label{sec:training_loss}

Training mediates between anchoring metric scale to sparse LiDAR and faithfully preserving the foundation model's intrinsic surface geometry:
\begin{equation}
    \mathcal{L}=\lambda_{\mathrm{depth}}\mathcal{L}_{\mathrm{depth}}+\lambda_{\mathrm{norm}}\mathcal{L}_{\mathrm{norm}}+\lambda_{\mathrm{scale}}\mathcal{L}_{\mathrm{scale}},
\end{equation}
where $\lambda_{\mathrm{depth}},\lambda_{\mathrm{norm}},\lambda_{\mathrm{scale}}$ are scalar weights balancing the three distinct loss terms.
\paragraph{Sparse depth alignment.}
$\mathcal{L}_{\mathrm{depth}}$ combines multi-resolution supervision and confidence-weighted regression:
\vspace{-6pt}
\begin{equation}
    \mathcal{L}_{\mathrm{sp}}\!=\!\frac{1}{K}\sum_{k=1}^{K}\frac{1}{|\mathcal{V}_k'|}\sum_{p\in\mathcal{V}_k'}\!\left|\hat{D}^{c,(k)}_p - D^{\mathrm{sp},(k)}_p\right|,
\end{equation}
% where $K$ is the number of supervision resolutions, $k$ indexes a resolution level, $p$ is a pixel index, $\hat{D}^{c,(k)}_p$ and $D^{\mathrm{sp},(k)}_p$ denote the predicted depth (from $\hat{\mathbf{D}}^{c}$) and projected sparse LiDAR depth at level $k$ and pixel $p$, $\mathcal{V}_k$ is the set of valid LiDAR pixels at level $k$, and $\mathcal{V}_k'\subseteq\mathcal{V}_k$ retains only the smallest 80\% of per-pixel errors to exclude misaligned projections.
where $K$ denotes the number of supervision resolutions and $k$ indexes each resolution level. For a given pixel $p$, $\hat{D}^{c,(k)}_p$ and $D^{\mathrm{sp},(k)}_p$ represent the predicted depth from $\hat{\mathbf{D}}^{c}$ and the projected sparse LiDAR depth at level $k$, respectively.
$\mathcal{V}_k$ is the set of valid LiDAR pixels at level $k$. Following MetricAnything~\cite{metricanything}, its subset $\mathcal{V}_k'\subseteq\mathcal{V}_k$ keeps only pixels with the smallest 80\% per-pixel errors, so as to filter out misaligned projections.

The confidence-weighted loss~\cite{dust3r} is formulated as:
\vspace{-4pt}
\begin{equation}
    \mathcal{L}_{\mathrm{conf}}\!=\!\frac{1}{|\mathcal{V}|}\sum_{p\in\mathcal{V}}\!\bigl(C_p|\hat{D}^{c}_p - D^{\mathrm{sp}}_p| - \lambda_c\log C_p\bigr),
\end{equation}
% \vspace{-2pt}
which enables adaptive downweighting of unreliable projections, where $\mathcal{V}$ is the set of valid LiDAR pixels at the native resolution, $C_p\in(0,1)$ is the predicted confidence at pixel $p$ taken from $\mathbf{C}$, and $\lambda_c$ balances the confidence regularization term to prevent trivial all-zero solutions.
The full term is $\mathcal{L}_{\mathrm{depth}}=\mathcal{L}_{\mathrm{sp}}+\mathcal{L}_{\mathrm{conf}}$.

\vspace{-5pt}
\paragraph{Surface-normal regularization.}
$\mathcal{L}_{\mathrm{norm}}$ penalizes angular deviation between normals from the corrected depth and those from the DA3 prior:
\vspace{-4pt}
\begin{equation}
    \mathcal{L}_{\mathrm{norm}}\!=\!\frac{1}{|\mathcal{M}|}\sum_{p\in\mathcal{M}}\!\bigl(1-\langle\mathcal{N}(\hat{\mathbf{D}}^{c}, \mathbf{K})_p,\,\mathcal{N}(\mathbf{D}^{\mathrm{prior}}, \mathbf{K})_p\rangle\bigr),
\end{equation}
% \vspace{-2pt}
where $\mathbf{K}$ is the camera intrinsics matrix, $\mathcal{N}(\mathbf{D},\mathbf{K})$ unprojects $\mathbf{D}$ into 3D using $\mathbf{K}$ and computes per-pixel surface normals via finite differences, $\langle\cdot,\cdot\rangle$ is the inner product between unit normals at pixel $p$, and $\mathcal{M}$ is the set of valid pixels excluding sky regions.
This enforces that neighboring pixels shift coherently, maintaining surface continuity rather than producing isolated spikes at LiDAR locations.

\paragraph{Scale regularization.}
In log-scale space $\mathbf{O}=\log\mathbf{S}_{\mathrm{pix}}$, with $O_p$ its value at pixel $p$ and $\partial_x,\partial_y$ horizontal and vertical finite differences:
\vspace{-2pt}
\begin{equation}
    \mathcal{L}_{\mathrm{scale}}\!=\!\frac{1}{|\mathcal{M}|}\sum_{p}\bigl(|O_p|+|\partial_x O_p|+|\partial_y O_p|\bigr),
\end{equation}
% \vspace{-pt}\noindent
% \vspace{-3pt}
penalizing both deviation from unity and spatial roughness between sparse anchors.

%======================================================================
\section{Experiments}
%======================================================================

We conduct experiments on nuScenes~\cite{nuscenes} and DDAD~\cite{ddad} to validate that DrivingDepth resolves the geometry--scale conflict: achieving metric calibration from sparse LiDAR prompts while preserving the foundation model's image-aligned geometry.

\subsection{Implementation Details}
\label{sec:impl_details}

% We initialize the foundation model from the  DA3NESTED-GIANT-LARGE-1.1~\cite{depthanything3} checkpoint.
We initialize the foundation model with the official DA3NESTED-GIANT-LARGE-1.1 checkpoint released in the DA3 repository~\cite{depthanything3}.
All experiments use 8 NVIDIA H20 GPUs with batch size 2 per GPU.
We train for 25 epochs over about 4 days, decaying the learning rate from $1\times10^{-4}$ to $1\times10^{-6}$.
Maximum depth is capped at 80\,m for both datasets, covering the effective distance required by downstream perception tasks.
Following DA3~\cite{depthanything3}, all input images are resized to $504\times280$.
For monocular baselines, we apply ROE~\cite{moge1} alignment to recover metric scale before evaluation.
For multi-frame methods, we compute a single global ROE scale shared across all cameras to preserve cross-view consistency.

\subsection{Datasets}
\label{sec:datasets}

\textbf{nuScenes v1.0}~\cite{nuscenes} follows the official split: 700 training scenes (28,130 samples) and 150 validation scenes (6,019 samples).
\textbf{DDAD}~\cite{ddad} contains 150 training scenes (12,650 frames) and 50 validation scenes (3,950 frames) with long-range driving sequences.
We use images, LiDAR measurements, camera intrinsics and extrinsics to construct sparse projected depth prompts and evaluation masks.
During training, we adopt mixed stride and mixed interval sampling with multiple candidate values to enrich input diversity.
Each input clip consists of $4$ frames captured across all $6$ surround-view camera views.

\subsection{Evaluation Metrics}
\label{sec:eval_metrics}

We report six metrics evaluating both metric accuracy and image--depth geometric and boundary consistency.
Standard metrics include AbsRel ($\downarrow$), $\delta_1$ ($\uparrow$), and $\delta_{0.5}$ ($\uparrow$).
We additionally report AbsIn02 ($\uparrow$), EdgeCR ($\uparrow$) and RevEdgeCR ($\uparrow$), defined as follows:

\begin{table}[h]
\centering
\small
\begin{tabular}{lc}
\toprule
Metric & Formula  \\
\midrule
AbsIn02 & $\frac{1}{N}\sum_{i=1}^{N}\mathbf{1}\!\left[|\hat{d}_i-d_i|<0.2\mathrm{m}\right]$ \\
EdgeCR & $\frac{\mathrm{mean}_{\mathcal{E}_I}(G_D)}{\mathrm{mean}_{\overline{\mathcal{E}}_I}(G_D)+\epsilon}$ \\
RevEdgeCR & $\frac{\mathrm{mean}_{\mathcal{E}_D}(G_I)}{\mathrm{mean}_{\overline{\mathcal{E}}_D}(G_I)+\epsilon}$ \\
\bottomrule
\end{tabular}
\caption{Additional evaluation metrics. $G_I{=}|\nabla I|$ and $G_D{=}|\nabla \hat{D}|$ are Sobel gradient magnitudes. Edge pixels are defined as $\mathcal{E}_I{=}\mathrm{Top}_{10\%}(G_I)$ and $\mathcal{E}_D{=}\mathrm{Top}_{10\%}(G_D)$.}
\label{tab:metrics}
\end{table}

AbsIn02 measures the fraction of pixels within 0.2\,m absolute error---directly reflecting whether predicted depth meets the spatial resolution of downstream occupancy grids.
EdgeCR checks whether RGB edges produce depth changes; RevEdgeCR checks whether depth edges are supported by visible image structures.
Together they quantify image--depth boundary alignment from both directions.

\begin{table*}[t]
\centering
\small
\begin{tabular}{lcccccccc}
\toprule
Method & Setting & LiDAR & AbsRel$\downarrow$ & $\delta_1$$\uparrow$ & $\delta_{0.5}$$\uparrow$ & AbsIn02$\uparrow$ & EdgeCR$\uparrow$  & RevEdgeCR$\uparrow$\\
\midrule
\multicolumn{9}{l}{\textit{Image-only foundation depth}} \\
Depth Pro~\cite{depthpro} & Mono & $\times$ & 16.86 & 76.47 & 62.90 & 29.93 & 5.404  & 2.381 \\
MOGE-2~\cite{moge2} & Mono & $\times$ & 13.78 & 83.64 & 72.25 & 37.97 & 5.226  & 2.393 \\
\midrule
\multicolumn{9}{l}{\textit{Sparse-prompted monocular depth}} \\
PromptDA~\cite{promptda} & Mono & \checkmark & 45.55 & 29.56 & 15.78 & 4.56 & 1.531  & 1.385 \\
PriorDA~\cite{prioranything} & Mono & \checkmark & 08.25 & 90.87 & 77.88 & 30.62 & 2.524  & 2.069 \\
\midrule
\multicolumn{9}{l}{\textit{Multi-view methods}} \\
DepthAnything3~\cite{depthanything3} & $4$F$\times6$V & $\times$ & 15.88 & 84.48 & 72.22 & 30.68 & \textbf{7.745}  & \textbf{2.597} \\
MapAnything~\cite{mapanything} & $4$F$\times6$V & \checkmark & 11.99 & \textbf{92.10} & \textbf{86.88} & 50.43 & 1.914  & 1.561 \\
\emph{DrivingDepth}\ (Ours) & $4$F$\times6$V & \checkmark & \textbf{11.19} & 89.96 & 85.22 & \textbf{61.14} & 5.741  & 2.273 \\
\bottomrule
\end{tabular}
\vspace{-5pt}
\caption{Comparison with representative methods on nuScenes~\cite{nuscenes}.  All approaches take $504\times280$ images as input and are post-processed with ROE~\cite{moge1} alignment. All sparse-prompted methods adopt the same projected LiDAR.
Best multi-view results are indicated in \textbf{bold}.}
\label{tab:main}
\end{table*}
\vspace{-10pt}

\begin{table}[t]
\centering

\label{tab:ten_percent_prompt_vertical_stack}
\setlength{\tabcolsep}{4pt} % 调整列间距，适配单栏宽度
\resizebox{0.8\linewidth}{!}{% 自适应单栏宽度，无内容溢出
\begin{tabular}{ll|c|cc|cc}

\toprule
% 表头第一行：所有方法
& & DA3 & MA & Ours & MA & Ours \\
% 表头第二行：对应LiDAR百分比
& LiDAR & -- & 10\% & 10\% & 100\% & 100\% \\
\midrule
% nuScenes Dataset 块，纵向堆叠第一部分
\multirow{5}{*}{\rotatebox[origin=c]{90}{\textbf{nuScenes}}} 
& AbsRel$\downarrow$ & 15.88 & 13.49 & \textbf{12.01} & 11.99 & \textbf{11.19} \\
& $\delta_1$$\uparrow$ & 84.48 & \textbf{89.97} & 89.22 & \textbf{92.10} & 89.96 \\
& $\delta_{0.5}$$\uparrow$ & 72.22 & 81.87 & \textbf{83.52} & \textbf{86.88} & 85.22 \\
& AbsIn02$\uparrow$ & 30.68 & 35.11 & \textbf{54.66} & 50.43 & \textbf{61.14} \\
& EdgeCR$\uparrow$ & 7.745 & 1.908 & \textbf{5.757} & 1.914 & \textbf{5.741} \\
\midrule % 两个数据集的分隔线
% DDAD Dataset 块，纵向堆叠第二部分
\multirow{5}{*}{\rotatebox[origin=c]{90}{\textbf{DDAD}}} 
& AbsRel$\downarrow$ & 14.32 & 09.91 & \textbf{08.59} & 09.56 & \textbf{07.80} \\
& $\delta_1$$\uparrow$ & 82.14 & \textbf{92.16} & 91.63 & \textbf{93.02} & 92.35 \\
& $\delta_{0.5}$$\uparrow$ & 63.01 & 85.13 & \textbf{85.35} & 83.97 & \textbf{87.15} \\
& AbsIn02$\uparrow$ & 18.84 & 33.23 & \textbf{39.71} & 30.47 & \textbf{44.99} \\
& EdgeCR$\uparrow$ & 13.17 & 1.787 & \textbf{9.146} & 1.766 & \textbf{9.110} \\
\bottomrule
\end{tabular}
}
\caption{Sparse-prompt robustness evaluation on nuScenes~\cite{nuscenes} and DDAD~\cite{ddad}, comparing model performance under 10\% and 100\% LiDAR point inputs. Here DA3 denotes DepthAnything3~\cite{depthanything3}, and MA stands for MapAnything~\cite{mapanything}. 
% All quantitative results are evaluated with full LiDAR observations for fair comparison.
}
\label{tab:ten_percent_prompt}
\vspace{-10pt}
\end{table}

\subsection{Main Results}
\label{sec:main_results}

\begin{figure*}[t]
  \centering
  \includegraphics[width=\textwidth]{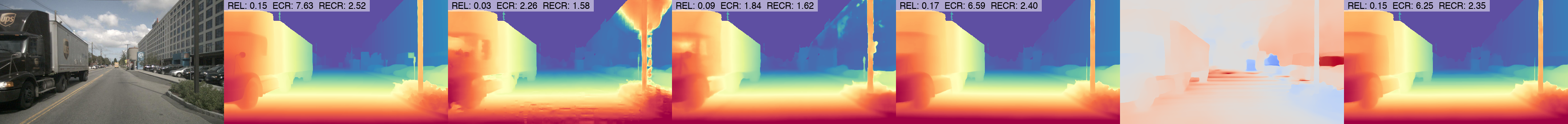}\\[1pt]
  \includegraphics[width=\textwidth]{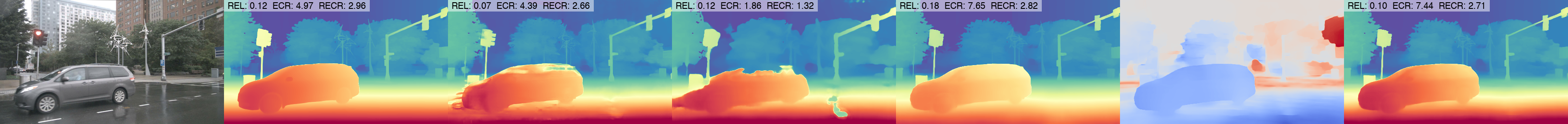}\\[1pt]
  \includegraphics[width=\textwidth]{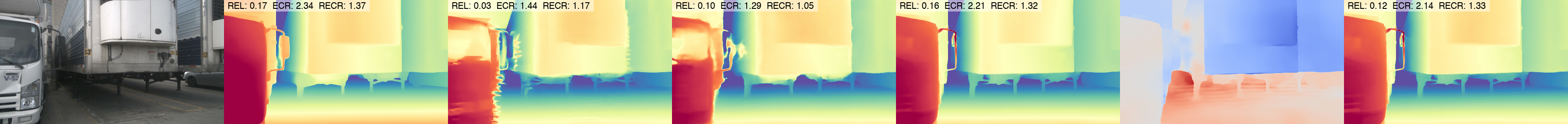}\\[1pt]
  \includegraphics[width=\textwidth]{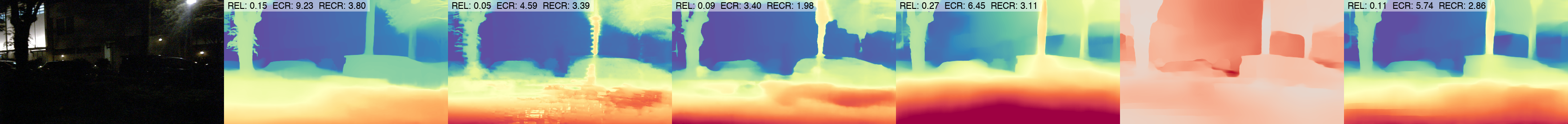}\\[1pt]
  \includegraphics[width=\textwidth]{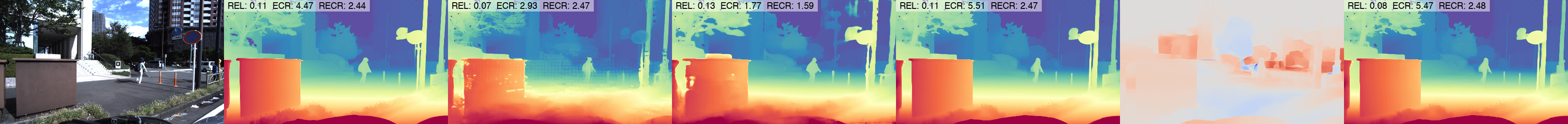}\\[1pt]
  \includegraphics[width=\textwidth]{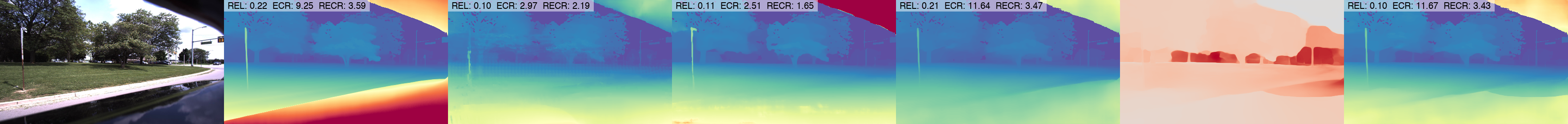}\\[0pt]
  \includegraphics[width=\textwidth]{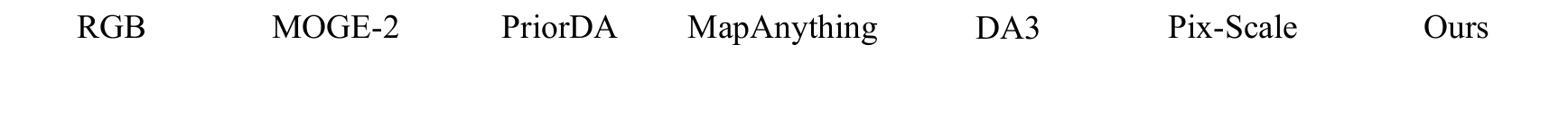}
    
  \vspace{-35pt}
  \caption{\textbf{2D depth comparison on nuScenes~\cite{nuscenes} (rows 1--4) and DDAD~\cite{ddad} (rows 5--6).} Columns: RGB, MOGE-2~\cite{moge2}, PriorDA~\cite{prioranything}, MapAnything~\cite{mapanything}, DA3~\cite{depthanything3}, $\mathbf{S}_{\mathrm{pix}}$, and DrivingDepth output. 
  Monocular methods use frame-wise alignment, while multi-view methods adopt clip-wise alignment.
  In $\mathbf{S}_{\mathrm{pix}}$: gray${\approx}1$, blue${<}1$, red${>}1$. DrivingDepth output approximates DA3 depth scaled element-wise by $\mathbf{S}_{\mathrm{pix}}$.
  % Monocular methods use frame-wise alignment, while multi-view methods adopt clip-wise alignment.
  }
  \label{fig:2d_comparison}
    \vspace{-13pt}

\end{figure*}

% We organize results around four axes: comparison with state-of-the-art methods, robustness to reduced LiDAR density, temporal and cross-view consistency, and sensitivity to input context.
We organize results around three axes: comparison with state-of-the-art methods, robustness to reduced LiDAR density, and sensitivity to input context.
Unless otherwise specified, the default uses 4 frames with interval 1.
\vspace{-5pt}
\paragraph{Comparison with state-of-the-art.}
Table~\ref{tab:main} groups methods by input setting.
Image-only foundation models achieve high EdgeCR from strong visual priors but limited metric accuracy without scale reference---precisely the geometry--scale conflict motivating this work.
Among sparse-prompted monocular methods, PriorDA~\cite{prioranything} obtains the lowest AbsRel by directly fitting LiDAR projections, but its EdgeCR drops to 2.524, indicating substantial deviation from image structures.
As visible in Figure~\ref{fig:2d_comparison}, PriorDA hallucinates depth discontinuities at LiDAR locations without corresponding image structure, confirming that naively fitting sparse points overrides foundation geometry.
PromptDA~\cite{promptda} fails entirely, likely because its architecture assumes denser prompts than driving LiDAR provides.
% In the multi-view setting, MapAnything achieves strong metric accuracy but sacrifices boundary alignment (EdgeCR 1.914), confirming that end-to-end approaches override foundation geometry.
In the multi-view setting, DepthAnything3~\cite{depthanything3} delivers promising boundary consistency and leverages multi-view inputs, yet it suffers from considerable depth estimation errors. 
MapAnything~\cite{mapanything} achieves strong metric accuracy but sacrifices boundary alignment (EdgeCR 1.914). This shows that the model rigidly fits LiDAR points and ignores the geometric structure of images.

DrivingDepth achieves competitive metric accuracy
% (AbsRel 11.19) 
while preserving significantly stronger boundary consistency (EdgeCR 5.741 vs.\ 1.914 for MapAnything~\cite{mapanything}), and leads on AbsIn02 by over 10 points---confirming that residual scale correction calibrates metric depth without overriding the frozen foundation's surface geometry.
\vspace{-5pt}
\paragraph{Sparse-prompt robustness.}
% Table~\ref{tab:ten_percent_prompt} evaluates performance when only 10\% of LiDAR points are retained.
% On nuScenes, DrivingDepth with 10\% prompts already surpasses MapAnything with full 100\% prompts on AbsIn02, while maintaining EdgeCR above 5.7.
% MapAnything degrades sharply: its AbsIn02 drops by over 15 points, while DrivingDepth drops by only 6.
% The same pattern holds on DDAD, where DrivingDepth at 10\% outperforms MapAnything at 100\% on metric accuracy while preserving EdgeCR above 9.
% This validates the multi-resolution prompt encoding and confidence-weighted loss: by treating LiDAR as sparse geometric prompts rather than dense supervision, the model extracts useful signal from extremely sparse inputs and automatically downweights unreliable points.
Table~\ref{tab:ten_percent_prompt} evaluates performance when only 10\% of LiDAR points are retained.
The DDAD~\cite{ddad} dataset contains denser and more accurate LiDAR points than nuScenes~\cite{nuscenes}, so methods can gain further performance improvements with richer and cleaner point observations.
On nuScenes, DrivingDepth with 10\% prompts already surpasses MapAnything~\cite{mapanything} with full 100\% prompts on AbsIn02, while maintaining EdgeCR above 5.7.
MapAnything degrades sharply: its AbsIn02 drops by over 15 points, while DrivingDepth drops by only 6.
The same pattern holds on DDAD, where DrivingDepth at 10\% outperforms MapAnything at 100\% on metric accuracy while preserving EdgeCR above 9.
Notably, several metrics of MapAnything even decline as LiDAR points increase on DDAD, revealing its poor robustness against dense noisy points.
This is caused by its tendency to rigidly fit LiDAR signals: although it achieves competitive $\delta$ metrics, its depth predictions fail to align with image semantics.
This validates the multi-resolution prompt encoding and confidence-weighted loss: by treating LiDAR as sparse geometric prompts rather than dense supervision, the model extracts useful signal from extremely sparse inputs and automatically downweights unreliable points.

\paragraph{Input-context analysis.}
Table~\ref{tab:context} evaluates generalization across frame counts and temporal spacings.
Single-frame inference---without cross-frame information---still achieves strong AbsRel (11.86), confirming that pixel-wise scale correction is effective on its own.
% Multi-frame input provides a 10-point boost on AbsIn02, showing that the feature adapter contributes complementary gains by propagating sparse-depth cues across frames when available.
Multi-frame input brings a 10-point boost on AbsIn02, demonstrating that multi-frame modeling achieves favorable performance under strict evaluation criteria.
Performance remains stable across 4--6 frames and intervals of 1--2, confirming robustness to temporal context variation.

\begin{table}[h]
\centering
\small
\begin{tabular}{ccccc}
\toprule
Frames & Interval & AbsRel$\downarrow$ & $\delta_1$$\uparrow$ & AbsIn02$\uparrow$ \\
\midrule
4 & 1  & 11.19 & 89.96 & 61.14 \\
6 & 1  & 11.27 & 90.02 & 61.02 \\
4 & 2  & 11.37 & 89.93 & 60.21 \\
1 & -- & 11.86 & 88.56 & 49.76 \\
\bottomrule
\end{tabular}
\caption{Input-context analysis varying frame count and temporal spacing on nuScenes~\cite{nuscenes}.}
  \vspace{-10pt}

\label{tab:context}
\end{table}

\subsection{Qualitative Results}
%======================================================================

\begin{figure*}[t]
  \centering
  \includegraphics[width=\textwidth]{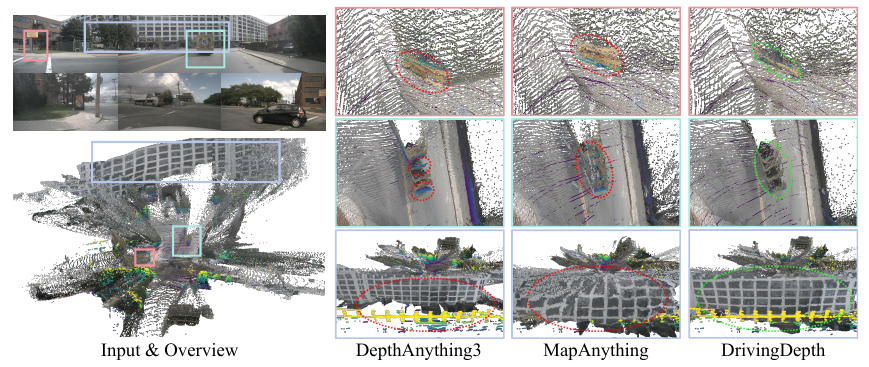}
    \vspace{-20pt}
  \caption{\textbf{3D reconstruction comparison on nuScenes~\cite{nuscenes}.} Predicted point clouds (RGB-colored) and LiDAR (colored by height) rendered together. Depth outputs of all methods are aligned using ROE\cite{moge1}. \textbf{Left}: RGB image (top) and DrivingDepth global point cloud (bottom). \textbf{Right}: zoomed crops comparing DA3~\cite{depthanything3}, MapAnything~\cite{mapanything}, and DrivingDepth.}
    \vspace{-10pt}

  \label{fig:3d_comparison}
\end{figure*}

\paragraph{2D depth comparison.}
% Figure~\ref{fig:2d_comparison} reveals a clear trade-off among existing approaches.
% Image-only methods (MOGE-2~\cite{moge2}, DA3~\cite{depthanything3}) produce the sharpest boundaries but without reliable metric scale.
% PriorDA~\cite{prioranything} achieves high metric accuracy but hallucinates depth discontinuities at LiDAR locations without corresponding image structure---overfitting to projections regardless of RGB semantics.
% MapAnything~\cite{mapanything} is sensitive to noisy projections: in the rainy scene (row~2), noise-induced artifacts are clearly visible.
% Row~4 presents a nighttime scene where monocular estimation is extremely challenging; through pixel-wise scale correction guided by sparse prompts, DrivingDepth still estimates plausible depth, correctly distinguishing foreground from background.
% The scale map $\mathbf{S}_{\mathrm{pix}}$ (column~6) confirms that the learned correction aligns with object-level boundaries, verifying spatially varying scale discovery rather than uniform adjustment.

Figure~\ref{fig:2d_comparison} visualizes predicted depth within 200\,m for all methods, with a unified sky mask from DA3 applied to all outputs for fair comparison. The figure reveals a clear trade-off among existing approaches.
Image-only methods such as MOGE~\cite{moge1,moge2} and DA3~\cite{depthanything3} produce the sharpest depth boundaries---MOGE-2~\cite{moge2} even outperforms DA3~\cite{depthanything3} on complex mixed backgrounds such as car mirrors in row~3---but neither provides reliable metric scale, which is the core motivation of this work.
PriorDA~\cite{prioranything} achieves high metric accuracy by directly fitting LiDAR projections, but it frequently hallucinates depth at locations where no corresponding image structure exists, effectively overfitting to projected points regardless of RGB semantics.
MapAnything~\cite{mapanything} is sensitive to noisy projections: in the rainy scene of row~2, noise-induced artifacts are clearly visible, while other methods remain relatively unaffected.
Row~4 presents a nighttime scene where the RGB image is largely indistinct, making monocular depth estimation extremely challenging. Yet through pixel-wise scale correction guided by sparse LiDAR prompts, DrivingDepth still estimates plausible background depth, correctly distinguishing foreground from the distant background even when visual cues are severely degraded.
Our predicted scale map $\mathbf{S}_{\mathrm{pix}}$ in column~6 confirms that the learned correction aligns with object-level boundaries, verifying that the model discovers spatially varying scale errors rather than applying uniform adjustment---consistent with our motivation that metric errors vary spatially across surfaces and depth ranges.
The final output in column~7 combines correct metric scale with pixel-aligned boundaries, achieving the balance that neither image-only nor existing sparse-prompted methods attain.

\vspace{-10pt}
\paragraph{3D reconstruction comparison.}
% Figure~\ref{fig:3d_comparison} visualizes two scenes as colored point clouds.
% DA3~\cite{depthanything3} suffers from scale misalignment: distant structures separate from LiDAR, producing visible layering where predicted depth peels away from true surfaces.
% MapAnything~\cite{mapanything} reduces scale error but introduces geometric distortions: the moving truck (Scene~1, green crop) shows blurred boundaries, and the vehicle (Scene~2, red crop) is visibly deformed because cross-view inconsistency allows one view's sparse pixels to dominate.
% DrivingDepth produces sharper boundaries with less blurring, leveraging the strong geometric prior from DA3, and achieves more geometrically faithful depth across all crops.
Figure~\ref{fig:3d_comparison} compares the 3D reconstruction results of different methods on a nuScenes scene, including colored point clouds and zoomed close-up views.
% Figure~\ref{fig:3d_comparison} visualizes two nuScenes scenes as colored point clouds with zoomed 3D crops. 
The global 3D point cloud shown in the bottom-left of each scene is produced by DrivingDepth, demonstrating that our method generates geometrically coherent surround-view reconstructions.
DA3~\cite{depthanything3} consistently suffers from scale misalignment: distant structures separate from LiDAR reference points, producing visible layering where the predicted depth peels away from the true surface.
MapAnything~\cite{mapanything} reduces scale errors but introduces geometric distortions. The moving truck highlighted by the green crop and the building in blue crop show broken boundaries.
% ; in Scene~2, the Subaru marked by the red crop is visibly deformed because cross-view inconsistency allows one view's sparse pixels to dominate the prediction, overriding geometry from other views.
% MapAnything~\cite{mapanything} also overfits to LiDAR in semantically inconsistent ways as seen in the blue crop of Scene~2, producing geometry that contradicts the visible image structure.
DrivingDepth produces sharper boundaries with less blurring thanks to the strong geometric prior from DA3, and achieves more geometrically faithful depth than MapAnything~\cite{mapanything} across all crops. The constrained multi-frame attention further ensures consistent geometry across views. These qualitative results corroborate the quantitative finding that residual scale correction preserves the foundation model's geometry while achieving metric calibration.

\begin{figure*}[t]
  \centering
  \begin{minipage}[c]{0.22\linewidth}
    \centering
    \small RGB + LiDAR\\[2pt]
    \includegraphics[width=\linewidth]{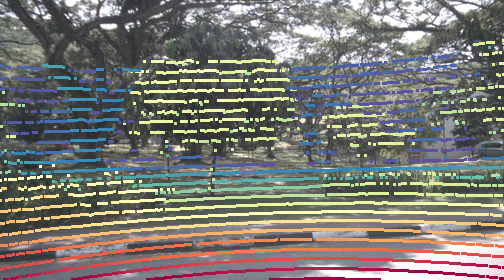}
  \end{minipage}%
  \hfill
  \begin{minipage}[c]{0.77\linewidth}
    \centering
    \setlength{\tabcolsep}{1pt}
    % \begin{tabular}{cccc}
    %   \small DA3& \small $\lambda_\text{norm}{=}10$ & \small $\lambda_\text{norm}{=}5$ & \small $\lambda_\text{norm}{=}2$ \\
    %   \includegraphics[width=0.24\linewidth]{norm_weight_imgs/depth_org.png}
    %   & \includegraphics[width=0.24\linewidth]{norm_weight_imgs/depth_surf10.png}
    %   & \includegraphics[width=0.24\linewidth]{norm_weight_imgs/depth_surf5.png}
    %   & \includegraphics[width=0.24\linewidth]{norm_weight_imgs/depth_surf2.png} \\[-2pt]
    %   \includegraphics[width=0.24\linewidth]{norm_weight_imgs/error_org.png}
    %   & \includegraphics[width=0.24\linewidth]{norm_weight_imgs/error_surf10.png}
    %   & \includegraphics[width=0.24\linewidth]{norm_weight_imgs/error_surf5.png}
    %   & \includegraphics[width=0.24\linewidth]{norm_weight_imgs/error_surf2.png} \\[-6pt]
    % \tiny rel=19.7 & \tiny rel=13.6 & \tiny rel=11.9 & \tiny rel=8.8 \\[-6pt]
    % \tiny edgecr=12.04 & \tiny edgecr=8.65 & \tiny  edgecr=8.52 & \tiny edgecr=7.78 \\[-6pt]
    % \tiny revedgecr=2.747 & \tiny revedgecr=2.426 & \tiny  revedgecr=2.425 & \tiny revedgecr=2.401 \\[-6pt]
    % \end{tabular}
    \newsavebox{\errimgbox}
    \sbox{\errimgbox}{\includegraphics[width=0.24\linewidth]{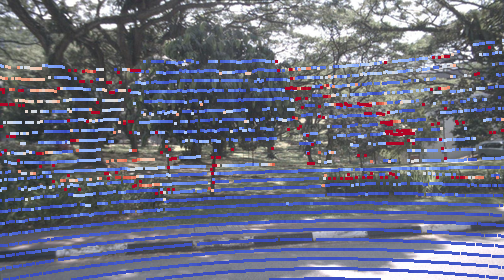}}
    
    \begin{tabular}{ccccc}
      \small DA3
      & \small $\lambda_\text{norm}{=}10$
      & \small $\lambda_\text{norm}{=}5$
      & \small $\lambda_\text{norm}{=}2$
      & \\
    
      \includegraphics[width=0.24\linewidth]{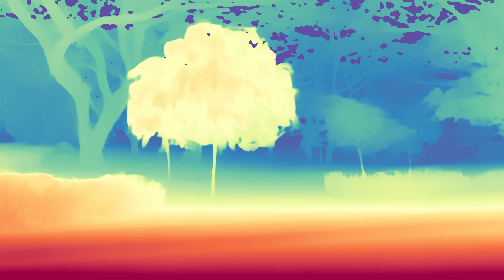}
      & \includegraphics[width=0.24\linewidth]{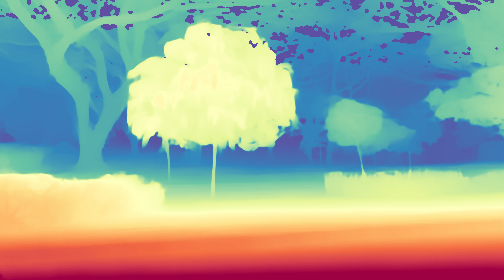}
      & \includegraphics[width=0.24\linewidth]{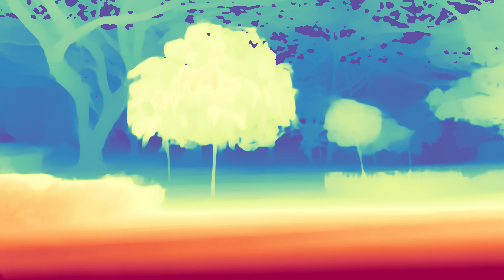}
      & \includegraphics[width=0.24\linewidth]{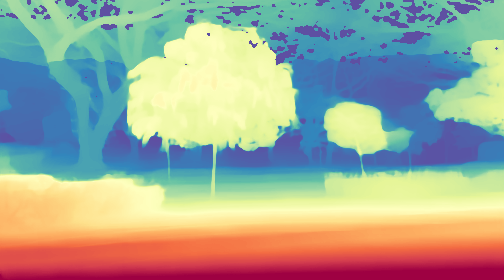}
      & \\[-2pt]
    
      \includegraphics[width=0.24\linewidth]{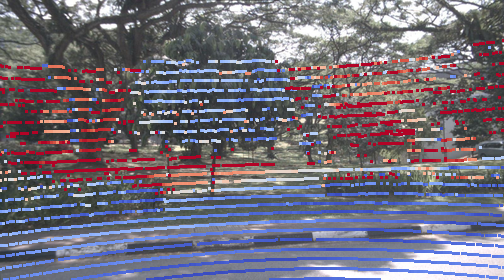}
      & \includegraphics[width=0.24\linewidth]{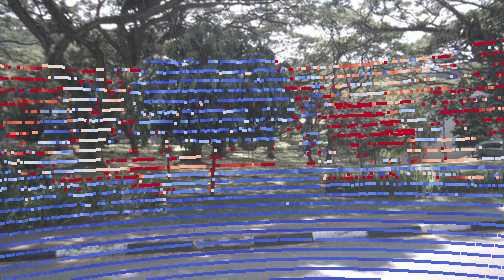}
      & \includegraphics[width=0.24\linewidth]{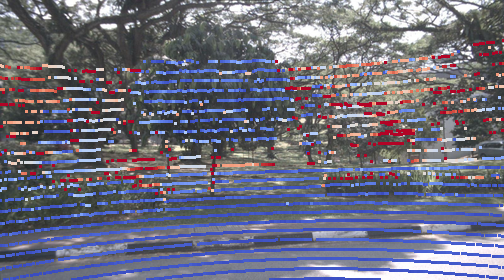}
      & \includegraphics[width=0.24\linewidth]{norm_weight_imgs/error_surf2.png}
      & \includegraphics[height=\ht\errimgbox]{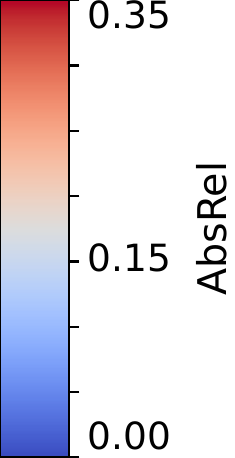} \\[-6pt]
    
      \scriptsize AbsRel=19.7 & \scriptsize AbsRel=13.6 & \scriptsize AbsRel=11.9 & \scriptsize AbsRel=8.8 & \\[-3pt]
      \scriptsize RevEdgeCR=2.747 & \scriptsize RevEdgeCR=2.426 & \scriptsize RevEdgeCR=2.425 & \scriptsize RevEdgeCR=2.401 &
    \end{tabular}
  \end{minipage}
  %
  % \begin{minipage}[t]{0.04\linewidth}
  %   \vspace{6pt}%
  %   \includegraphics[height=0.7cm]{norm_weight_imgs/colorbar_v.png}
  % \end{minipage}
  \vspace{-7pt}
  \caption{\textbf{Surface-normal weight trade-off visualization.} Left: RGB with sparse LiDAR overlay. Top row: predicted depth under different $\lambda_\text{norm}$. Bottom row: per-pixel absolute relative error (blue=low, red=high). At $\lambda_\text{norm}{=}2$, a visible horizontal line at the LiDAR coverage boundary indicates that the model breaks image-aligned structure in unsupervised regions when normal regularization is insufficient.}
  \label{fig:normal_weight}
    \vspace{-10pt}

\end{figure*}

%======================================================================
\subsection{Ablation Studies}
\label{sec:ablation}
%======================================================================

\begin{table}[h]
\centering
\small
\resizebox{\columnwidth}{!}{%
\setlength{\tabcolsep}{3pt}
\begin{tabular}{lccccc}
\toprule
    & Mul-Res & GPFA & AbsRel$\downarrow$ & $\delta_1$$\uparrow$ &  EdgeCR$\uparrow$  \\
\midrule
(a) Full model & \checkmark & \checkmark & \textbf{11.19} & \textbf{89.96}  & 5.741   \\
(b) w/ ~~CVA & \checkmark & \checkmark  & 11.38 & 89.56  & 5.771  \\
(c) w/o GPFA & \checkmark & $\times$  & 12.49 & 88.49  & \textbf{5.825}  \\
(d) w/o Mul-Res   & $\times$ & \checkmark & 13.13 & 87.80  & 5.787 \\
(e) w/o LiDAR     &  -- &  $\times$& 13.39 & 87.46  & 5.776 \\
% (f) Raw DA3    &  -- &   --  & 15.88 & 84.48  & 7.745 \\
\bottomrule
\end{tabular}%
}
\caption{Component ablation on nuScenes~\cite{nuscenes}. CVA denotes the cross-view attention within the same frame. GPFA denotes the Geometry-Preserving Feature Adapter.}
\vspace{-10pt}
\label{tab:component}
\end{table}

\paragraph{Component ablation.}
Table~\ref{tab:component} quantifies the individual contribution of each core module.
Replacing the constrained cross-view and cross-frame attention within GPFA with vanilla cross-view attention (b) yields mild performance degradation. Vanilla attention blindly establishes correlations between all camera pairs, bringing geometric noise from unrelated views even though cross-view modeling partially offsets this loss.
Fully removing the entire GPFA module (c) triggers a much larger decline in AbsRel and $\delta_1$. Without GPFA, the model cannot inject sparse depth cues into intermediate backbone layers nor strengthen feature exchange between spatially overlapping camera views.
Discarding multi-resolution prompt (d) further degrades model performance, with metrics drawing near the LiDAR-free baseline. This trend corroborates our earlier analysis: single-resolution projected driving LiDAR is extremely sparse, leaving thinly sampled regions without sufficient prompt signals and introducing severe high-frequency noise into decoder feature maps.
% When all LiDAR input is removed entirely (e), the model forfeits all metric calibration signals and falls back solely on the intrinsic depth prior of the frozen DA3 base model, achieving the worst overall depth accuracy among all ablation configurations.
When all LiDAR input is completely removed (e), the model discards all external metric calibration cues and merely relies on the intrinsic depth prior of frozen DA3, corresponding to a lightweight fine-tuning over the original DA3 pipeline. It delivers the weakest results among all variants with LiDAR guidance.
Across every tested variant, EdgeCR only fluctuates moderately between 5.75 and 5.83. This stable boundary metric confirms our geometric constraints consistently maintain sharp object boundaries regardless of the metric calibration modules enabled.
Taken together, these ablation results consistently validate the effectiveness of each component in our overall pipeline design.

\begin{table}[h]
\centering
\small
\begin{tabular}{ccccc}
\toprule
$\lambda_{\mathrm{norm}}$ & AbsRel$\downarrow$ & $\delta_1$$\uparrow$ & AbsIn02$\uparrow$ &  EdgeCR$\uparrow$  \\
\midrule
2 & 10.16 & 91.38 & 62.64 & 5.667 \\
5 & 11.19 & 89.96 & 61.14 & 5.741  \\
10 & 11.79 & 89.27 & 58.63 & 5.776 \\
\bottomrule
\end{tabular}
% \caption{Surface-normal regularization weight ablation on nuScenes~\cite{nuscenes}.}
\caption{Ablation study for surface-normal regularization weight \(\lambda_{\mathrm{norm}}\) on nuScenes~\cite{nuscenes}.}
\label{tab:normal_weight}
  \vspace{-10pt}

\end{table}

\vspace{-20pt}
\paragraph{Surface-normal regularization trade-off.}
Table~\ref{tab:normal_weight} and Figure~\ref{fig:normal_weight} study the balance between metric accuracy and surface plausibility.
Weaker regularization ($\lambda_{\mathrm{norm}}{=}2$) allows more aggressive LiDAR fitting, yielding the best AbsRel but lower EdgeCR---the correction introduces structural artifacts between anchors.
As shown in Figure~\ref{fig:normal_weight}, at $\lambda_{\mathrm{norm}}{=}2$ a visible horizontal line appears at the LiDAR coverage boundary, revealing that without sufficient normal regularization the model destroys image-aligned structure in unsupervised regions.
Stronger regularization $\lambda_{\mathrm{norm}}{=}10$ enforces smoother surfaces but over-constrains scale correction, degrading AbsRel by 0.6 points.
We select $\lambda_{\mathrm{norm}}{=}5$ as the operating point balancing metric accuracy against surface coherence.
This smooth trade-off directly reflects the geometry--scale conflict: the surface-normal loss provides a continuous control knob between the two competing objectives, validating our problem decomposition.

%======================================================================
\section{Conclusion}
%======================================================================

This paper addresses the inherent geometry--scale conflict in driving depth estimation by proposing DrivingDepth, a lightweight sparse-prompted framework that calibrates foundation depth through residual pixel-wise scale correction.
The key insight is that foundation models already provide geometrically coherent relative depth; what remains is only the per-pixel scale mapping this geometry to metric coordinates.
By keeping the backbone frozen and learning only a residual correction initialized at unity, the method preserves dense visual geometry by construction while achieving metric calibration from sparse LiDAR prompts.
Extensive experiments on nuScenes~\cite{nuscenes} and DDAD~\cite{ddad} demonstrate competitive metric accuracy with significantly stronger boundary consistency than existing methods, and robust performance under 10\% LiDAR density.
A current limitation is the reliance on a single frozen foundation model; if its geometry is fundamentally incorrect (e.g., on reflective or transparent surfaces), the scale correction cannot recover the missing structure.

%======================================================================
%======================================================================

% \clearpage % Force references to start on a new page
{
    \small
    \bibliographystyle{ieeenat_fullname}
    \bibliography{main}
}
\end{document}